\pdfoutput=1

\documentclass[11pt]{article}

\usepackage{coling}

\usepackage{times}
\usepackage{latexsym}

\usepackage[T1]{fontenc}

\usepackage[utf8]{inputenc}

\usepackage{microtype}

\usepackage{inconsolata}

\usepackage{graphicx}

\usepackage{latexsym}
\usepackage{amsmath}
\usepackage{amsthm}
\usepackage{booktabs} 
\usepackage{multirow} 
\usepackage{hyperref}
\usepackage{subcaption}

\title{Building a Family of Data Augmentation Models for Low-cost LLM Fine-tuning on the Cloud}

\author{Yuanhao Yue$^{1,2}$\thanks{Work done during the internship at Alibaba Cloud Computing.}, Chengyu Wang$^{2}$\thanks{Corresponding authors.}, Jun Huang$^{2}$, Peng Wang$^{1}$\footnotemark[2]\\
  $^{1}$ School of Computer Science, Fudan University, Shanghai, China\\
  $^{2}$ Alibaba Cloud Computing, Hangzhou, China\\
  \texttt{yhyue22@m.fudan.edu.cn}\\
  \texttt{\{chengyu.wcy,huangjun.hj\}@alibaba-inc.com}\\
  \texttt{pengwang5@fudan.edu.cn}
  }

\begin{document}
\maketitle
\begin{abstract}
Specializing LLMs in various domain-specific tasks has emerged as a critical step towards achieving high performance. However, the construction and annotation of datasets in specific domains are always very costly. Apart from using superior and expensive closed-source LLM APIs to construct datasets, some open-source models have become strong enough to handle dataset construction in many scenarios. Thus, we present a family of data augmentation models designed to significantly improve the efficiency for model fine-tuning. These models, trained based on sufficiently small LLMs, support key functionalities with low inference costs: instruction expansion, instruction refinement, and instruction-response pair expansion. To fulfill this goal, we first construct an automatic data collection system with seed datasets generated from both public repositories and our in-house datasets. This system leverages powerful LLMs to expand, refine and re-write the instructions and responses, incorporating quality assessment techniques. Following this, we introduce the training process of our models, which effectively distills task-solving and text synthesis abilities from teacher LLMs. Finally, we demonstrate how we integrate these functionalities into a machine learning platform to support low-cost LLM fine-tuning from both dataset preparation and training perspectives for users. Experiments and an application study prove the effectiveness of our approach.~\footnote{All the produced data augementation models have been released:~\href{https://huggingface.co/alibaba-pai/Qwen2-1.5B-Instruct-Exp}{Qwen2-1.5B-Instruct-Exp},~\href{https://huggingface.co/alibaba-pai/Qwen2-7B-Instruct-Exp}{Qwen2-7B-Instruct-Exp},~\href{https://huggingface.co/alibaba-pai/Qwen2-1.5B-Instruct-Refine}{Qwen2-1.5B-Instruct-Refine},~\href{https://huggingface.co/alibaba-pai/Qwen2-7B-Instruct-Refine}{Qwen2-7B-Instruct-Refine} and~\href{https://huggingface.co/alibaba-pai/Qwen2-7B-Instruct-Response-Exp}{Qwen2-7B-Instruct-Response-Exp}.}
\end{abstract}

\section{Introduction}

The advent of large language models (LLMs) has revolutionized the landscape of NLP, offering unprecedented capabilities in understanding and generating human language~\cite{DBLP:journals/tist/ChangWWWYZCYWWYZCYYX24,DBLP:journals/csur/MinRSVNSAHR24}.
However, for industrial practitioners, fine-tuning LLMs is crucial to solve tasks that may not be adequately addressed by existing LLMs.

Previous studies illustrate that LLMs fine-tuned with calibrated datasets can surpass those trained on larger, but quality-compromised datasets~\cite{DBLP:conf/nips/ZhouLX0SMMEYYZG23,DBLP:journals/corr/abs-2308-12032}. However, assembling high-quality datasets is expensive, tedious and time-consuming, often putting state-of-the-art techniques out of reach for many developers and industrial practitioners, due to the ``data hunger'' problem. 
Data augmentation strategies, such as paraphrasing, have been proposed to bolster the volume of training data~\cite{DBLP:conf/acl/AbaskohiRY23,DBLP:conf/acl/ZhouZTJY22}. 
These functionalities are critical for enterprise clients operating in cloud environment.
However, for LLMs, the challenge of data augmentation becomes paramount. It not only involves expanding the volume of datasets but also enhancing the clarity and precision of instructions, and fostering enriched instruction-response pairs.

In this paper, we introduce a family of data augmentation models to reduce the dependency on large volumes of high-quality instructional data for LLM fine-tuning, which empower users with functionalities such as instruction expansion, refinement, and the generation of enriched instruction-response pairs with minimal inference costs. Our approach involves an automatic data collection system that synthesizes seed datasets from both public repositories and our proprietary datasets. This system harnesses the capabilities of powerful LLMs to incrementally polish and regenerate textual data, with quality assessment to ensure the utility of augmented datasets. By embedding our models into a cloud-native machine learning platform, we enable practical, low-cost fine-tuning that substantially reduces the burdens of dataset preparation and model training.
Experiments and an application study show the efficacy of our approach.

\section{Related Work}

In this section, we briefly overview of the related work on LLMs and data augmentation.

\subsection{Large Language Models}
Prior to the surge of LLMs, Pre-trained Language Models (PLMs) had captivated widespread interest due to their proficiency in acquiring contextualized representations~\cite{DBLP:journals/corr/abs-2003-08271}. 
A typical example is BERT~\cite{DBLP:conf/naacl/DevlinCLT19}, which leverages the encoder-only design, which has found wide application across various language comprehension tasks. With the advent of ChatGPT, there has been an influx of diverse LLMs introduced to the field. Notable among these publicly accessible LLMs are the LLaMA series~\cite{DBLP:journals/corr/abs-2302-13971,DBLP:journals/corr/abs-2307-09288}, the Qwen series~\cite{DBLP:journals/corr/abs-2309-16609}, OPT~\cite{DBLP:journals/corr/abs-2205-01068}, Galactica~\cite{DBLP:journals/corr/abs-2211-09085}, GLM~\cite{DBLP:conf/acl/DuQLDQY022}, among others. A key step for LLMs to follow human instructions is instruction tuning (or called supervised fine-tuning), proposed by~\citet{DBLP:conf/iclr/WeiBZGYLDDL22} and followed by a variety of works~\cite{DBLP:journals/corr/abs-2308-10792}.
Our work on data augmentation is orthogonal to the aforementioned studies, signifying that it can enhance the effectiveness of instruction tuning for any LLM backbones. Due to space limitation, we do not elaborate.

\subsection{Data Augmentation}
Data augmentation is the process of artificially expanding a dataset by generating new data points from existing ones. This is done through various transformations that alter the data while still maintaining its core properties. For text data, traditional augmentation techniques involve synonym replacement, word insertion or swapping, back-translation, or sentence shuffling~\cite{DBLP:conf/acl/FengGWCVMH21}. Recently, several strategies, such as paraphrasing and textual entailment, have been proposed to augment the data from the semantic level~\cite{DBLP:conf/acl/AbaskohiRY23,DBLP:conf/acl/ZhouZTJY22,DBLP:conf/emnlp/Kumar0SZ22}.
For LLMs, data augmentation is usually applied to the prompt level for better instruction tuning, i.e, the generation of more instructions, responses or instruction-response pairs. For example,~\citet{DBLP:conf/acl/WuZH23} leverage chain-of-thought prompting to augment knowledge for reasoning tasks.~\citet{DBLP:conf/acl/ZhouLJB23} propose dual prompt augmentation for cross-lingual tasks. PromptMix~\cite{DBLP:conf/emnlp/SahuVBL23} generates augmented data by utilizing LLMs to perform few-shot classification tasks. In contrast to previous works, our trained models exhibits versatility and can be deployed across a diverse range of NLP tasks based on the instruction tuning paradigm.

\begin{figure}[t]
\centering
\includegraphics[width=0.5\textwidth]{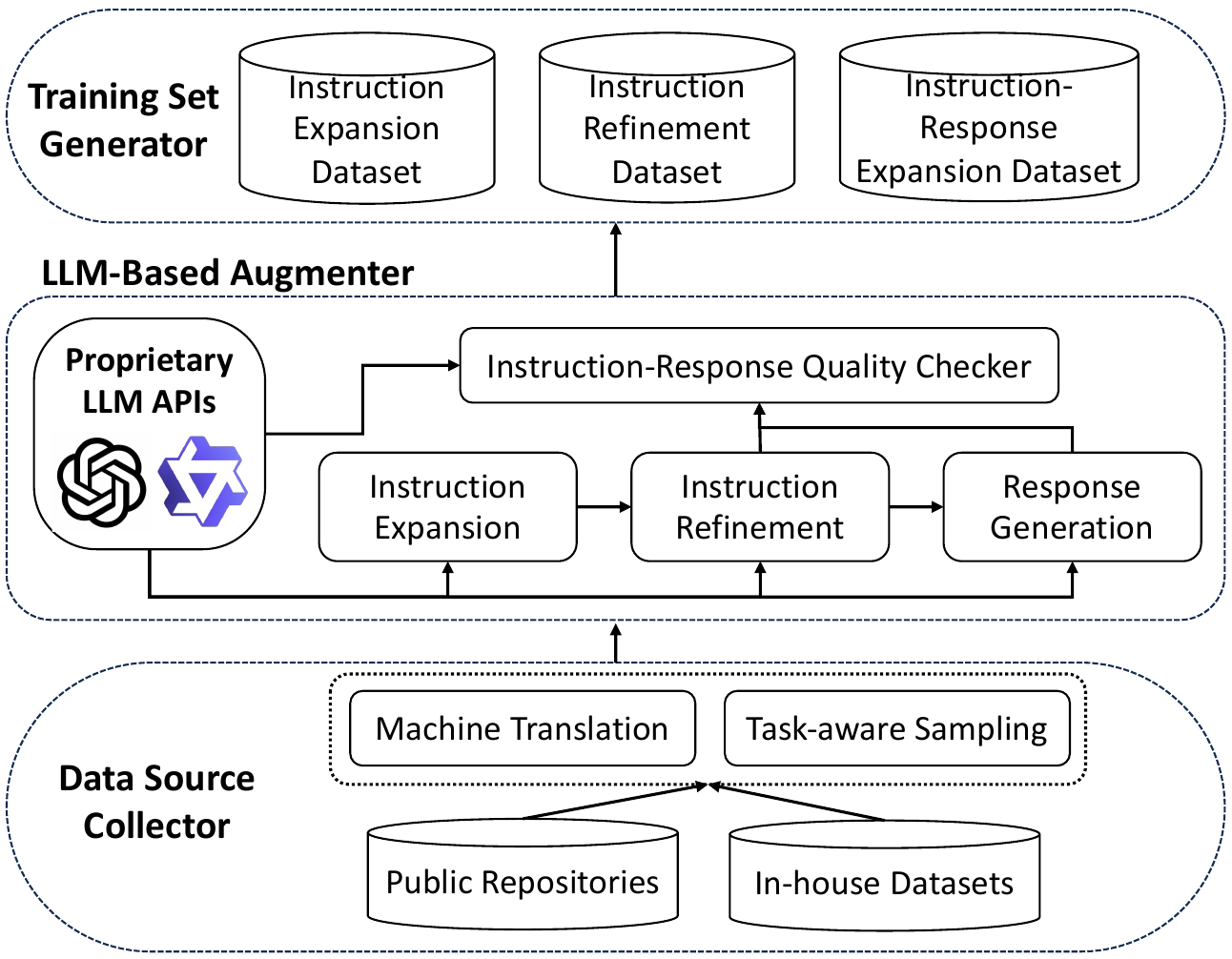}
\caption{The data collection system. 
} 
\label{datac-collect}
\end{figure}

\begin{table*}[t]
\centering
\begin{scriptsize}
\begin{tabular}{ll}
\toprule
\bf Original & You are an expert in Transformer models. How to implement a Transformer model using PyTorch?\\
\midrule
\bf Refined & As an expert in Transformer models, please provide a detailed guide explaining how to implement a Transformer model using the PyTorch\\
&  framework. Please include the following key sections: \\
& 1. Model Architecture Overview: Describe the basic structure and components of the Transformer.\\
& 2. Implementation Steps: Detail the specific steps to implement the Transformer using PyTorch, including defining the model, configuring\\
&  layers, writing the forward propagation function, etc.\\
& 3. Parameter Settings: Provide recommended settings for key parameters such as learning rate, batch size, etc.\\
& 4. Training Process: Explain how to train the model, including how to prepare the data, choose the loss function and optimizer, etc.\\
& 5. Model Evaluation: Describe how to evaluate the model's performance, including how to perform validation and testing.\\
& Please ensure the guidance is accurate and detailed to facilitate understanding and application by beginners.\\
\toprule
\bf Original & Create a travel guide for Hangzhou.\\ 
\midrule
\bf Refined & Create a comprehensive Hangzhou travel guide containing key information. The guide should include:\\
& 1. Introduction and recommended itinerary for major attractions in Hangzhou.\\
& 2. Recommended local foods and restaurant information.\\
& 3. Accommodation suggestions, including options for different budget levels.\\
& 4. Local transportation guide, including how to get from the airport to the city center and recommended transportation between attractions.\\
& 5. Visitor tips, such as the best travel seasons, local cultural etiquette, etc.\\
& Based on the above requirements, please create a complete Hangzhou travel guide.\\
\bottomrule
\end{tabular}
\end{scriptsize}
\caption{Examples of how we re-writes user's prompts sampled from the LLM service.}
\label{tab_refine_example}
\end{table*}

\section{The Proposed Approach}

In this section, we present our work on data augmentation models for low-cost LLM fine-tuning.

\subsection{Data Collection System}

The high-level architecture of our data collection system is shown in Figure~\ref{datac-collect}. The system consists of three major modules introduced below.

\subsubsection{Data Source Collector}
This module aims to generate a sufficiently large, diverse (in types of NLP tasks) and high-quality~\emph{seed dataset}, consisting of instruction-response pairs, as the input to our   system. As reported in~\citet{DBLP:conf/nips/ZhouLX0SMMEYYZG23}, the diversity and quality of instructional data are vital to the effectiveness of instruction tuning.
Here, we combine several public datasets including OpenHermes 2.5\footnote{https://huggingface.co/datasets/teknium/OpenHermes-2.5}, Cleaned Alpaca Dataset\footnote{https://github.com/gururise/AlpacaDataCleaned} and LCCD~\cite{DBLP:conf/nlpcc/WangKZHJZH20}, together with the in-house dataset sampled from LLM online API services to capture the preference of online users.
As we mostly focus on the English and Chinese languages in our cloud service, we also leverage machine translation systems to translate all the collected instruction-response pairs into the two languages if not present. The source data collection process for other languages can be conducted in a similar fashion.

To balance the task distributions of instructional data, an important step is~\emph{task-aware sampling}~\cite{DBLP:journals/corr/abs-2405-13448}. We conduct re-sampling of the collected pairs to create a more task-balanced seed dataset. Finally, we finish compiling our dataset, containing 36K instruction-response pairs.

\subsubsection{LLM-Based Augmenter}

It is important to point out that the goal of our trained models is not~\emph{generating good responses to instructions}, but specializing~\emph{augmenting instructional data on user demand}. In this module, we leverage powerful, proprietary LLMs to synthesize augmentation data. Here, we employ~\emph{Qwen-max}\footnote{https://qwenlm.github.io/} for augmenting texts in Chinese (which has better abilities for the Chinese language), and~\emph{GPT-4} for others. Three sub-tasks are defined as follows.

\noindent\textbf{Instruction Expansion.}
The task is to expand current instruction pool by generating instructions with similar task types but different targets, compared to seed ones as in-context demonstrations. For example, given a seed instruction ``\emph{Plan an in-depth tour itinerary of France that includes Paris, Lyon, and Provence.}'', possible outputs include:
\begin{enumerate}
    \item \emph{Describe a classic road trip itinerary along the California coastline in the United States.}
    \item  \emph{Create a holiday plan that combines cultural experiences in Bangkok, Thailand, with beach relaxation in Phuket.}
\end{enumerate}

\noindent\textbf{Instruction Refinement.}
The writing and style of instructions are crucial for effectively conversing with LLMs, commonly known as~\emph{prompt engineering}~\cite{DBLP:journals/corr/abs-2302-11382}. In the literature, instruction refinement is often leveraged to guide LLMs to generate better responses for specific tasks~\cite{DBLP:conf/emnlp/ShumDZ23,DBLP:conf/iclr/0001Z0S23}. Here, we ask powerful LLMs to act as a skilled prompt engineer to refine the instructions in our dataset. We demonstrate how prompt refinement works in Table \ref{tab_refine_example}.
The generated refined instructions can significantly prompt LLMs to produce better and more informative responses for users.

\noindent\textbf{Response Generation.}
With expanded and refined instructions, 
we manually annotated several examples to write an in-context learning prompt (see Table \ref{tab:refining_prompt}) to ask these powerful LLMs to generate responses with higher quality and more details. This step is similar to distill the knowledge from these LLMs for training specialized small models \cite{DBLP:journals/corr/abs-2405-13448,DBLP:conf/acl/HsiehLYNFRKLP23}.

In addition, to ensure the generated instructions and instruction-response pairs are factually correct, we leverage the LLMs to check the data quality and filter out low-quality ones.  The prompt templates for instruction expansion, refinement and quality checking are listed in Appendix \ref{ap:prompt_for_all}.

\begin{table}[t]
\centering
\begin{small}
\begin{tabular}{llllll}
\toprule
\bf Statistics & $I_{src}$ & $I_{tgt}$ & $I_{tgt}^{(*)}$ & $I$ & $R$\\ 
\midrule
$\mathcal{D}_{IE}$ & 10K & - & 20K & - & -\\
$\mathcal{D}_{IR}$ & 36K & 36K & - & - & -\\
$\mathcal{D}_{IRE}$ & - & - & - & 20K & 20K\\
\bottomrule
\end{tabular}
\end{small}
\caption{Statistics of the generated datasets.}
\label{dataset}
\end{table}

\subsubsection{Training Set Generator}

After the augmentation process, we obtain the following three training sets for fine-tuning our models, with statistics summarized in Table~\ref{dataset}.
i) The instruction expansion dataset $\mathcal{D}_{IE}$ consists of the tuples of a source and several target instructions $\mathcal{I}_{IE}=(I_{src},I_{tgt}^{(1)},I_{tgt}^{(2)},\cdots,I_{tgt}^{(N)})$ where $I_{tgt}^{(*)}$ is expanded from $I_{src}$ and $N$ is the number of generated samples for a source instruction.
ii) The instruction refinement dataset $\mathcal{D}_{IR}$ consists of source and target instruction pairs $(I_{src},I_{tgt})$, where $I_{tgt}$ is refined from $I_{src}$. 
iii) The instruction-response expansion dataset $\mathcal{D}_{IRE}$ consists of instruction-response pairs $(I,R)$. Its annotations come from $D_{IE}$. We use \emph{Qwen-max} to annotate responses for all the instructions in $D_{IE}$, and construct the training set in the form of Table \ref{tab:train_exp_response_prompt}, using the expanded annotations of one of instructions in the in-context examples as the output. In order to increase the diversity of the training pairs generated by the model after fine-tuning, we randomly shuffle 15\% of the model output annotations.

Note that different from $\mathcal{D}_{IE}$ and $\mathcal{D}_{IR}$ where instructions in a data sample are strongly co-related in terms of task types, $\mathcal{D}_{IRE}$ can be viewed as an enlarged and quality-improved version of our original seed dataset. Thus, our functionality of instruction-response expansion allows the free generation of any new instruction-response pairs, which will be elaborated in the next part.

\subsection{Model Training}

We first introduce the training loss of our models. For cloud service, we wish to lower the batch inference costs for users as much as possible. Therefore, specialized small models that excel in one task are more desirable.
Denote $\Phi$ as the collection of parameters of the underlying LLM for each task. For~\emph{instruction expansion} (IE), we define the loss function $\mathcal{L}_{IE}$, shown as follows:
\begin{equation}
\mathcal{L}_{IE}=-\sum_{\mathcal{I}_{IE}\in\mathcal{D}_{IE}}\sum_{i}^{N}\log\Pr(I_{tgt}^{(i)} | I_{src}; \Phi)
\end{equation}
which considers multiple expanded instructions for each source instruction $I_{src}$. 

For~\emph{instruction refinement} (IR), the loss function $\mathcal{L}_{IR}$ is more straightforwardly formulated, which follows the widely-used causal auto-regressive language modeling process, formulated as follows:
\begin{equation}
\mathcal{L}_{IR}=-\sum_{(I_{src},I_{tgt})\in\mathcal{D}_{IR}}\log\Pr(I_{tgt} | I_{src}; \Phi).
\end{equation}

Finally, for the~\emph{instruction-response expansion} (IRE) task, we seek to produce a relatively more powerful LLM than those for IE and IR that is capable of generating new instruction-response pairs. Based on our enterprise-level requirements, these pairs are not required to share the same task type with that of user input. Hence, given $K$ input pairs as seed user dataset, our model requires to output new ones using the $K$ pairs as in-context demonstrations. Let $(I_i,R_i)\in \mathcal{D}_{IRE}$ be a target sample, and $(I_i^{(1)},R_i^{(1)}),(I_i^{(2)},R_i^{(2)}),\cdots,(I_i^{(K)},R_i^{(K)})\in \mathcal{D}_{IRE}$ be $K$ randomly sampled in-context samples that are not overlapped with $(I_i,R_i)$
The loss function of the task $\mathcal{L}_{IRE}$ is defined as follows:
\begin{equation}
\begin{split}
\mathcal{L}_{IRE}= &-\sum_{(I_i,R_i)\in\mathcal{D}_{IRE}} \log\Pr(I_i,R_i |I_i^{(1)},R_i^{(1)},\\
&I_i^{(2)},R_i^{(2)},\cdots,I_i^{(K)},R_i^{(K)}; \Phi).
\end{split}
\end{equation}
During training of the three types of models, we carefully craft user prompts and system prompts, with templates detailed in Appendix \ref{ap:prompt_for_all}.  

As for model backbones, we leverage the chat models of the Qwen2 series~\cite{DBLP:journals/corr/abs-2309-16609} for further fine-tuning. The reasons for our choice are twofold.
i) It provides pre-trained models in various parameter scales. ii) Compared to other model series, it has good mastery in both English and Chinese, which are our major target languages.
We choose backbones that best fit our tasks and keep the models as small as possible to reduce inference costs.
The produced final model list, together with the key information, can be found in Table~\ref{model}.

\begin{table}[t]
\centering
\begin{small}
\begin{tabular}{ll}
\toprule
\bf Function & \bf Model\\ 
\midrule
IE & \emph{Qwen2-1.5B-Instruct-Exp}\\
IE & \emph{Qwen2-7B-Instruct-Exp}\\
\midrule
IR & \emph{Qwen2-1.5B-Instruct-Refine}\\
IR & \emph{Qwen2-7B-Instruct-Refine}\\
\midrule
IRE & \emph{Qwen2-7B-Instruct-Response-Exp}\\
\bottomrule
\end{tabular}
\end{small}
\caption{The model list. We do not train IRE models on 1.5B scale as such small models lack capacity to write high-quality and diverse instruction-response pairs.}
\label{model}
\end{table}

\begin{figure}[t]
\centering
\includegraphics[width=0.5\textwidth]{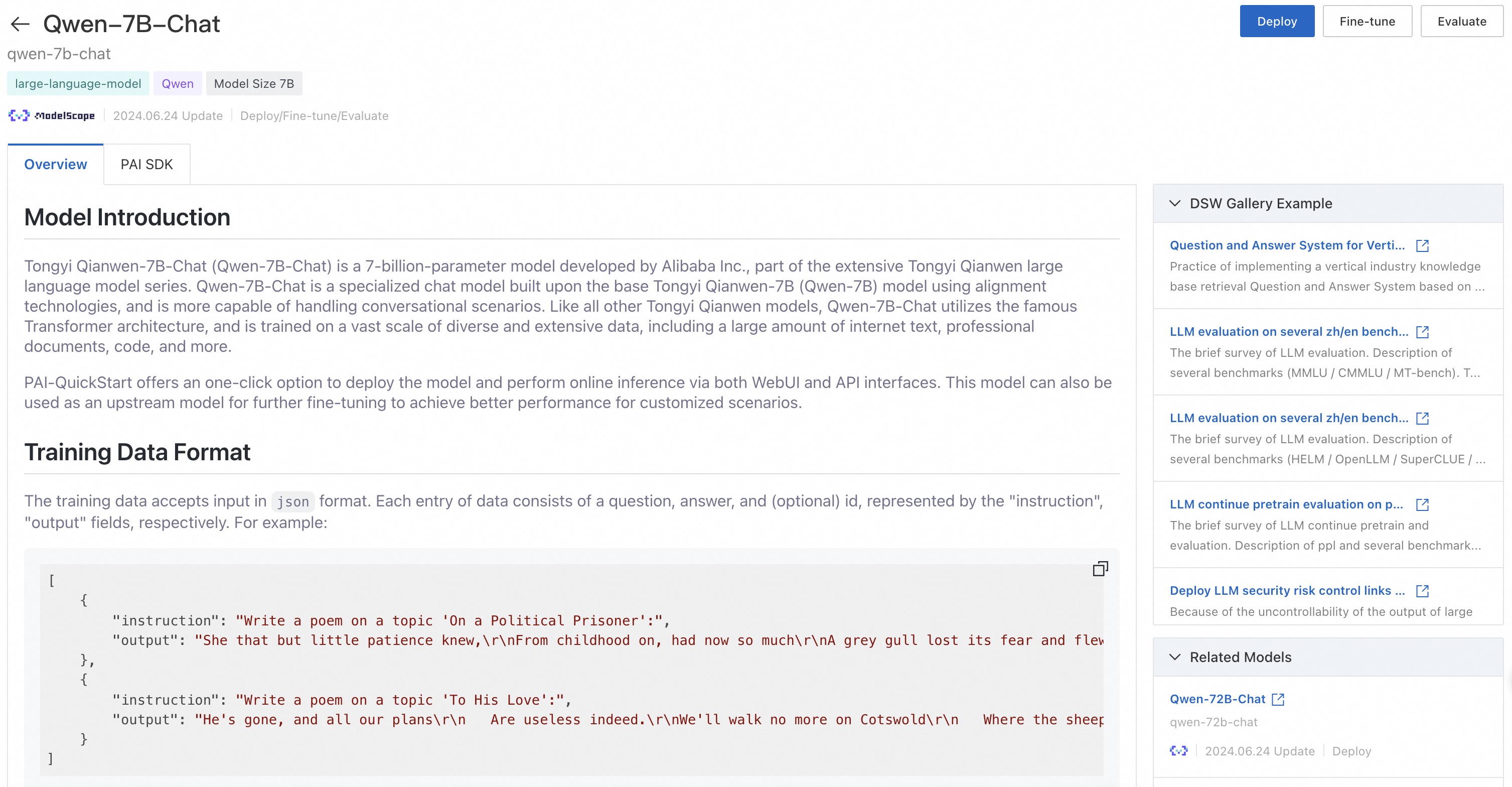}
\caption{A snapshot of the model card.} 
\label{card}
\end{figure}

\subsection{Integration to Cloud-native Machine Learning Platform}

Apart from release of our trained data augmentation models to the open-source community, we have integrated the data augmentation functionalities to a cloud-native machine learning platform (Alibaba Cloud Platform For AI) to facilitate low-cost LLM fine-tuning from both perspectives of data preparation and training strategies.

Given a~\emph{seed user dataset}, a~\emph{data pipeline} begins by augmenting the number of instructions by the IE model, with responses automatically distilled by a user-specified off-the-shelf LLM. Users also have the liberty to provide ground-truth responses to new instructions themselves.
Next, two optional steps can be conducted on demand, including re-writing the instructions using the IR models, and augmenting the entire dataset using the IRE model.

The~\emph{training pipeline} supports various types of LLM algorithms, including standard fine-tuning, RLHF~\cite{DBLP:conf/nips/Ouyang0JAWMZASR22}, DPO~\cite{DBLP:conf/nips/RafailovSMMEF23}, etc. To save the GPU memory consumption, several parameter-efficient training strategies can be applied to these algorithms with ease, e.g., LoRA~\cite{DBLP:conf/iclr/HuSWALWWC22}, QLoRA~\cite{DBLP:conf/nips/DettmersPHZ23}, etc, which is not the major focus of this work.
A snapshot of one of our model cards is shown in Figure~\ref{card}. Readers can also refer to our application studies for more examples.

\begin{table}[t]
\centering
\begin{small}
\begin{tabular}{llll}
\toprule
\bf Model & \bf Math   & \bf  Impl. \\ 
\midrule
Qwen2-1.5B-Instruct& 57.90\%  &28.96\% \\
\midrule
\emph{+ Qwen2-1.5B-Instruct-Exp} & 59.15\%  &31.22\% \\
\emph{+ Qwen2-7B-Instruct-Exp} & 58.32\%  & 39.37\% \\
\midrule
\midrule
Qwen2-7B-Instruct & 71.40\%  & 28.85\% \\
\midrule
\emph{+ Qwen2-1.5B-Instruct-Exp} & 73.90\% & 35.41\%\\
\emph{+ Qwen2-7B-Instruct-Exp} & 72.53\%  & 32.92\% \\
\bottomrule
\end{tabular}
\end{small}
\caption{Effectiveness of IE models on two challenging tasks.}
\label{eval1}
\end{table}

\begin{table}[t]
\centering
\begin{small}
\begin{tabular}{lcc}
\toprule
\bf Model & \bf Detail & \bf Truthfulness\\ 
\midrule
Qwen2-1.5B-Instruct & 50.00\% & 50.00\%\\
\midrule
\emph{+ Qwen2-1.5B-Instruct-Refine} & 75.63\% & 63.75\%\\
\emph{+ Qwen2-7B-Instruct-Refine} & 76.56\% &62.19\% \\
\midrule
\midrule
Qwen2-7B-Instruct & 50.00\% & 50.00\%\\
\midrule
\emph{+ Qwen2-1.5B-Instruct-Refine} & 70.94\% & 57.19\%\\
\emph{+ Qwen2-7B-Instruct-Refine} & 74.69\% &58.44\% \\
\bottomrule
\end{tabular}
\end{small}
\caption{The relative win rate of our IR models in terms of level of details and truthfulness relative to original instructions with two different response LLMs.}
\label{eval2}
\end{table}

\begin{table}[t]
\centering
\begin{small}
\begin{tabular}{lcccc}
\toprule
\bf Diversity & \bf Length & \bf Complexity & \bf Factuality\\ 
\midrule
\multicolumn{4}{c}{\centering Self-Instruct} \\
  9.6 & 15.8 & 0.32 & 5.0 \\
\midrule
\multicolumn{4}{c}{\centering \emph{Qwen2-7B-Instruct-Response-Exp}} \\
17.2 & 26.3 & 4.97 & 4.9 \\
\bottomrule
\end{tabular}
\end{small}
\caption{Effectiveness of IRE models in four aspects, compared with Self-Instruct.}
\label{eval3}
\end{table}

\section{Experiments and Application Study}
In this section, we present the experimental results to verify the effectiveness of our approach. After that, we show how our models can be utilized to support real-world applications.
In the experiments, we train the models listed in Table~\ref{model} using our collected datasets. We train our model with a learning rate of $1\times10^{-5}$ for 3 epochs. All the experiments are conducted on a sever with A100 GPUs (80GB).
 

\subsection{Effectiveness of IE}

We evaluate our instruction expansion models on two tasks from the BIG-Bench benchmark \cite{big-bench}. We choose tasks spanning logical reasoning and commonsense. We split a subset of 100 data instances as seed dataset for the \textit{Implicature} dataset and 1000 data points for the \textit{Elementary Math} dataset. We employ our instruction expansion models to expand the seed data to six times its original size., and use \emph{Qwen-max} to annotate the newly generated data. From Table \ref{eval1}, we can observe that despite the \textit{Qwen2-Instruct} models having already undergone extensive training in the domain of mathematics, our data augmentation technique can still consistently improve the model's performance by an additional 1-2 percentage points. In contrast, for the \textit{Implicature} dataset where the model has not been extensively trained, data augmentation results in a more significant improvement in performance, with an increase of approximately 7-11 percentage points. We further visualize the instruction expansion in Figure \ref{expand-visulize} in the appendix.  

\subsection{Effectiveness of IR}

For IR evaluation, we take single-turn instructions from a widely-used benchmark MT-Bench~\cite{DBLP:conf/nips/ZhengC00WZL0LXZ23} as input to Qwen2-1.5B-Instruct and Qwen2-7B-Instruct to generate responses, which are regarded as the vanilla method with any refinement. Two IR models are further leveraged to refine these instructions, before response generation. After that, we employ~\emph{GPT4-turbo} to evaluate the levels of details and truthfulness of the responses, compared with the vanilla outcomes. The relative win rates of our IR models are shown in Table~\ref{eval2}, with results of our vanilla method set to be 50\%.
From the results, we can see that our IR models consistently improve the response quality over multiple response LLMs in two aspects. Particularly, the improvement over the smaller 1.5B model is more significant, because smaller LLMs have weaker task-solving capacities, and hence require detailed instructions to deliver good responses.

\subsection{Effectiveness of IRE}

We follow the experimental procedures of Self-Instruct~\cite{self-instruct} , utilizing the same 175 human-written instructions as seeds to expand to 1,000 instructions. For comparison, we sample 1,000 entries from the Alpaca dataset expanded by Self-Instruct \cite{self-instruct}. We then compare the two dataset expansion methods in terms of data diversity, length, complexity, and factuality. We calculate the diversity of the dataset by counting the unique bigrams of the instruction per example. The average number of tokens of the instruction per example is used as the length value for each dataset. We use the perplexities obtained from LLaMA3-8B\footnote{https://huggingface.co/meta-llama/Meta-Llama-3-8B} to calculate the average IFD \cite{IFD} score for each dataset as an assessment of data complexity. Finally, we use~\emph{GPT4-turbo} to evaluate the factuality of the instruction-response pairs in the datasets. From Table \ref{eval3}, we can observe that as our model extends to datasets with higher complexity and diversity, its truthfulness approaches that of the Self-Instruct \cite{self-instruct}. We visualize the two datasets in Figure \ref{exp-resp-visualize}. Data expanded by \emph{Qwen2-7B-Instruct-Response-Exp} spans a more diverse range of regions within the embedding space, compared to the data expanded by Self-Instruct.

\begin{figure}[t]
\centering
\includegraphics[width=0.5\textwidth]{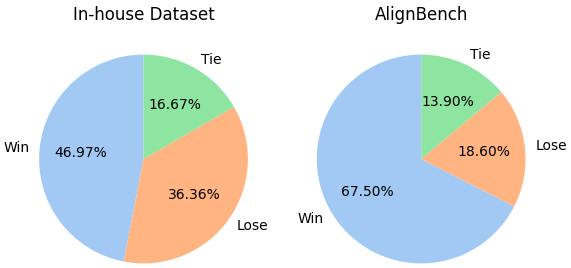}
\caption{The win-lose-tie rates of \emph{Qwen2-7B-Instruct-Refine} for the prompt refinement task, compared with the much larger model~\emph{Qwen-max}.} 
\label{win}
\end{figure}

\subsection{Application Studies}

We further show the efficacy of our approach in refining user prompts for LLM-based chatbots, which shows our work can be also beneficial for LLM inference scenarios, apart from fine-tuning.
 
It is common knowledge that instruction-tuned LLMs can naturally serve as chatbots; however, their effective use can be challenging for beginners without experiences to craft detailed and informative prompts. Therefore, LLMs are commonly employed as prompt engineers to enhance user experience. In a mobile chatbot application, the chat pipeline integrates a large proprietary LLM, i.e.,~\emph{Qwen-max} as the prompt engineer. As a result, two separate inference procedures (one for refinement and the other for response) are necessary to generate better responses when the refinement procedure is invoked. To address the challenge, our IR model (i.e.,~\emph{Qwen2-7B-Instruct-Refine}) can be utilized as a compact tool to refine user prompts.

We conduct a user study in which we randomly sample a collection of online user prompts, denoted as our in-house dataset, together with a public benchmark AlignBench~\cite{DBLP:journals/corr/abs-2311-18743} for instruction tuning evaluation in Chinese, and refine them using both the proprietary model and our~\emph{Qwen2-7B-Instruct-Refine}. 
The qualities of resulting prompts by both models are evaluated by~\emph{GPT-4-turbo}, and we report the rates of win-lose-tie (i.e., whether~\emph{Qwen2-7B-Instruct-Refine} beats~\emph{Qwen-max}), comparing the two prompt refinement models.  
The results, presented in Figure~\ref{win}, indicate that our model achieves comparable and sometimes better performance while significantly reducing the parameter size from several hundreds of billions to just 7B. 
Examples of some refined cases are illustrated in Table~\ref{tab_refine_example}, with texts translated from Chinese to English.
In the future, we seek to i) deploy the model online to reduce inference time and conserve computational resources for prompt refinement, and ii) provide offline batch inference service for users on the cloud.

\section{Conclusion}

In summary, our paper presents a novel and economical strategy for fine-tuning LLMs by introducing data augmentation models that decrease the necessary data for effective training. By utilizing smaller LLMs and an automatic data collection system, we offer a solution that reduces both computational and financial constraints. Experimental results and application studies confirm the efficiency of our approach, making LLMs more accessible for users with limited resources.

\section*{Limitations}

Despite the promising outcomes of our data augmentation models for fine-tuning LLMs, our approach is not without limitations. Firstly, the performance of our system is inherently tied to the quality and diversity of the initial seed datasets. If these datasets possess biases or are not representative of the target domain, the augmentation process might propagate or amplify these limitations. Secondly, while our system reduces the need for extensive datasets, there is still a dependency on publicly available LLMs. The quality and capabilities of these smaller LLMs can constrain the upper bound of effectiveness. Lastly, while the integration into a cloud-native platform suggests scalability, there might be operational challenges and costs associated with cloud computing that were not comprehensively assessed in our study. These limitations highlight the need for further research to enhance the robustness and applicability of data augmentation approaches in LLM fine-tuning.

\section*{Ethical Considerations}

While our approach seeks to democratize fine-tuning LLMs by data augmentation, it could inadvertently contribute to exacerbating existing biases in the data. Since our trained models rely on public datasets and LLMs, they are subject to the inherent biases present in these sources. If not carefully monitored, our system could perpetuate these biases through the generated instructions and responses, leading to unfair outcomes. Furthermore, the process could enable malicious actors to create language models for harmful purposes, such as generating fake news, spam, or other types of deceptive content. The implications of making such powerful technology more accessible necessitate careful consideration of safeguards and monitoring to prevent abuse.

\section*{Acknowledgments}

This work was supported by Alibaba Research Intern Program.


\appendix

\section{Visualization of Augmented Data Distributions}

\begin{figure*}[t]
\centering
\includegraphics[width=0.6\textwidth]{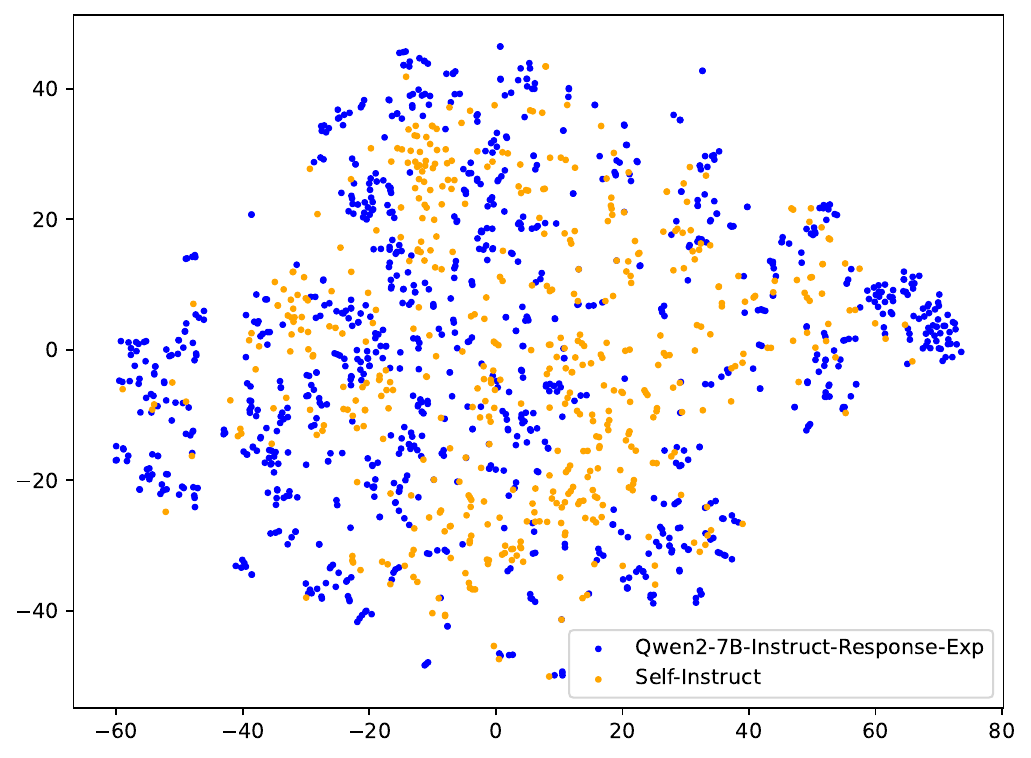}
\caption{We observe that the data generated by \emph{Qwen2-7B-Instruct-Response-Exp}, compared to data generated by Self-Instruct, occupies a more broadly distributed range of regions within the embedding space after being projected to two dimensions using t-SNE.} 
\label{exp-resp-visualize}
\end{figure*}

\begin{figure*}[h]
   \centering
   \begin{subfigure}{0.49\textwidth}
      \includegraphics[width=\linewidth]{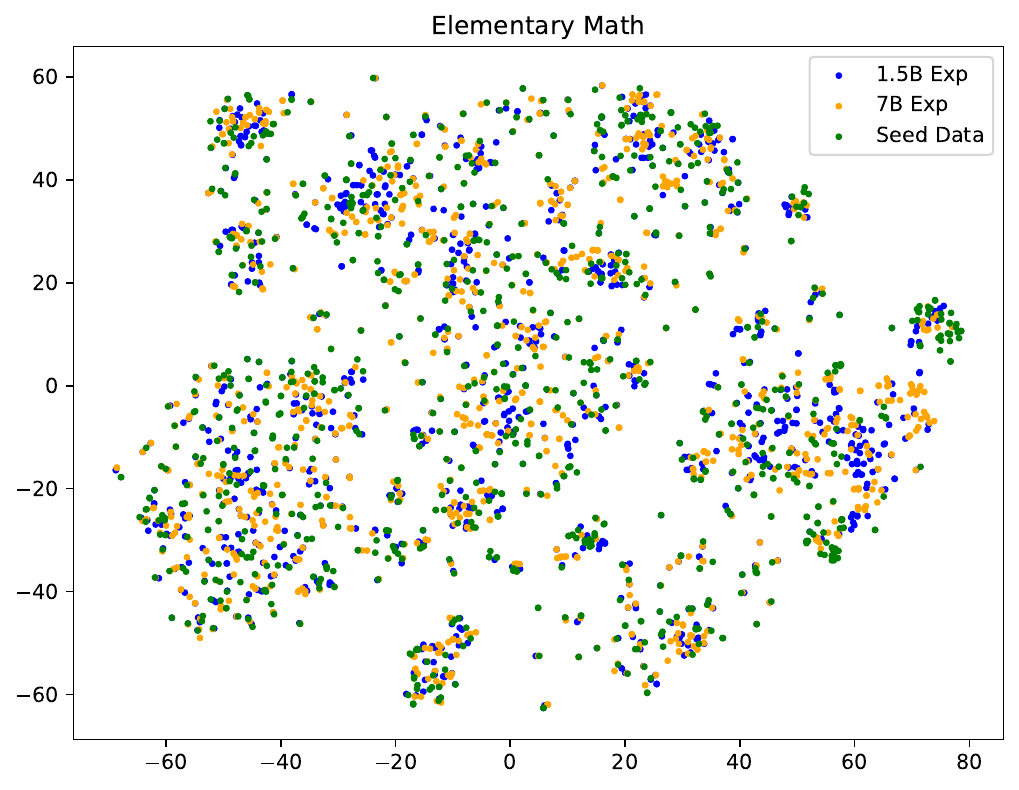}
      \caption{Visualization of t-SNE dimensionality reduction for the expanded data and the original seed data.}
   \end{subfigure}\hfill
   \begin{subfigure}{0.49\textwidth}
      \includegraphics[width=\linewidth]{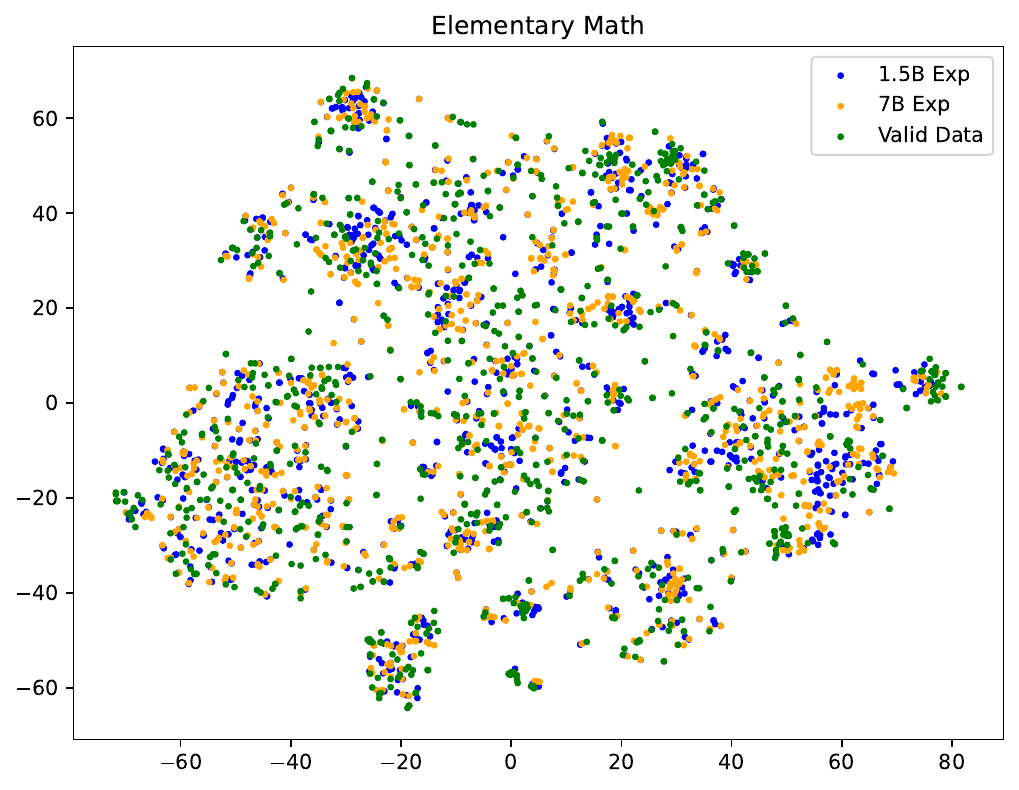}
      \caption{Visualization of t-SNE dimensionality reduction for the expanded data and the validation data.}
   \end{subfigure}
   \caption{Distribution of the model expansion and human-written dataset in the embedding space on the Elementary Math dataset. Datasets augmented by our models exhibit substantial regional overlap with the seed dataset, consequently leading to significant overlap with most regions of the validation set. The data generated by the \emph{Qwen2-7B-Instruct-Exp} is slightly smoother and more uniform compared to that produced by the \emph{Qwen2-1.5B-Instruct-Exp}.}
   \label{expand-visulize}
\end{figure*}

\section{Prompt Templates}
\label{ap:prompt_for_all}

\subsection{Prompt Templates for Generating Training Sets}

\begin{table*}[!h]
\small
\centering
\begin{tabular}{l}
\toprule
As a skilled prompt engineer, your expertise lies in refining prompts to be more efficient. Your task is to refine a given\\
user prompt, ensuring that the resulting prompt is clearer and more structured.\\ \\
The refined prompt must stay true to the user's original intent, possibly adding context or any information that narrows\\
down the scope and guides the large model for better understanding and task completion. The user's prompt should be\\
restructured with care to avoid excessive expansion.\\ \\
Essential details from the user's initial prompt, such as background knowledge relevant to the task, source text in text\\
analysis assignments, and requirements about the output format, must be preserved in the refined prompt.\\ \\
If the initial prompt is lengthy, consider inserting separators to make the structure of the refined prompt more visible.\\ \\
Should the user's prompt contain variables like "\$\{\{variable\_name\}\}", these must remain in the refined prompt. You\\ may introduce additional configurable variables, represented as "\$\{\{new\_variable\_name\}\}", to allow the prompt to\\
support further user-provided details.\\ \\
The language of the refined prompt should match that of the user's prompt. If the user's prompt is in Chinese, then\\
the refined prompt must also be in Chinese; similarly, if the user's prompt is in English, the refined prompt must also\\
be in English.\\ \\
Please output only the refined prompt without extraneous content, such as "\#\#Refined Prompt\#\#". 
\\ \\ Here are some examples:\\ \\
\#\#User's Prompt\#\#: \\ \\
Painting, music. Select the correct pairing for the given words.\\ \\
\#\#Refined Prompt\#\#: \\ \\
Choose an appropriate match for the terms "painting" and "music". \\ \\
\#\#User's Prompt\#\#: \\ \\
Analyze the structure of the following news article. \$\{\{news\}\} \\ \\
\#\#Refined Prompt\#\#: \\ \\
Analyze the headline and subtitle of the following news article, detailing how they establish the theme, capture\\
reader interest, and provide background context. Discuss how the specific choice of words and structure of the\\
headline and subtitle efficiently convey the central message of the news.\\
\$\{news\}\}  \\ \\
\#\#User's Prompt\#\#: \\ \\If a customer inquires about product specifications without specifying the product, prompt them for more details.\\
Answer fully using document content without excessive explanation.\\ \\
\#\#Refined Prompt\#\#: \\ \\Instruction: When answering customer inquiries about product specifications, if the customer does not mention\\
a specific product, request additional details from the customer. \\
Response Format: Use a formal and professional customer service tone to answer based on handbook information\\
regarding product specifications.\\
Considerations:\\
    1. If the customer does not specify product details, use this template to reply: "Hello! To provide accurate product\\
    specifications, could you please specify which product you're referring to?"\\
    2. Once the customer provides the details of a specific product, respond with accurate and comprehensive\\
    specification data.\\
    3. Avoid irrelevant explanations and ensure the response is concise, directly addressing the customer's queries.\\ \\
\#\#User's Prompt\#\#: \\ \\
\{prompt\_to\_refine\} \\ \\
\#\#Refined Prompt\#\#: \\

\bottomrule
\end{tabular}
\caption{Prompt template for annotating prompt refinement.}
\label{tab:refining_prompt}
\end{table*}

\begin{table*}[!h]
\small
\centering
\begin{tabular}{l|l}
\toprule
System prompt &
\parbox[c]{13cm}{
\texttt{You are a helpful assistant.}
}\\
\toprule
User prompt &
\parbox[c]{13cm}{
\texttt{} \\ 
\texttt{I want you to act as an Instruction Creator.} \\ 
\texttt{Your goal is to draw inspiration from the \#Given Instruction\# to create a brand new instruction. } \\ 
\texttt{ This new instruction should belong to the task type of [{task\_type}] as the \#Given Instruction\#.
} \\ 
\texttt{The LENGTH and difficulty level of the \#Created Instruction
\# should be similar to that of the \#Given Instruction\#.} \\
\texttt{The content of the \#Created Instruction\# should be different from that of the \#Given Instruction\#.} \\ 
\texttt{The \#Created Instruction\# must be reasonable and must be understood and responded to by humans.} \\
\texttt{’\#Given Instruction\#’, ’\#Created Instruction\#’, ’given instruction’ and ’created instruction’ are not allowed to appear in \#Created Instruction\#.} \\
\texttt{\#Given Instruction\#:} \\
\texttt{\{instruction\}} \\
\texttt{\#Created Instruction\#:}
} \\
\bottomrule
\end{tabular}
\caption{Prompt template for annotating dataset expansion.}
\label{tab:dataset_expansion}
\end{table*}

\subsection{Prompt Templates for Model Training}

\begin{table*}[!h]
\small
\centering
\begin{tabular}{l|l}
\toprule
System Prompt &
\parbox[c]{13cm}{
\texttt{You are a helpful assistant to refine this instruction and modify it into a more precise and detailed instruction.}
} \\
\toprule
User prompt & \texttt{\{instruction\_to\_refine\}}\\
\midrule
Model Output & \texttt{\{refined\_instruction\}}\\
\bottomrule
\end{tabular}
\caption{Prompt template for training instruction refining models.}
\label{tab:train_refine_prompt}
\end{table*}

\begin{table*}[!h]
\small
\centering
\begin{tabular}{l|l}
\toprule
System Prompt &
\parbox[c]{13cm}{
\texttt{You are a helpful assistant to expand this instruction to an instruction of the same task type but with different content.}
} \\
\toprule
User prompt & \texttt{\{instruction\_to\_expand\}}\\
\midrule
Model Output & \texttt{\{expanded\_instruction\}}\\
\bottomrule
\end{tabular}
\caption{Prompt template for training instruction expansion models.}
\label{tab:train_expantion_prompt}
\end{table*}

\begin{table*}[!h]
\small
\centering
\begin{tabular}{l|l}
\hline
System Prompt &
\parbox[c]{13cm}{
\texttt{You are a helpful assistant to continue writing the following instruction-response pairs.}
} \\
\hline
User Prompt &
\parbox[c]{13cm}{
\texttt{\#\#\# Instruction:} \\ 
\texttt{\{instruction\_1\}} \\
\texttt{\#\#\# Response:} \\
\texttt{\{response\_1\}} \\
\texttt{......} \\
\texttt{\#\#\# Instruction:} \\ 
\texttt{\{instruction\_n\}} \\
\texttt{\#\#\# Response:} \\ 
\texttt{\{response\_n\}} \\
\texttt{\#\#\# Instruction:} 
}\\
\hline
Model Output &
\parbox[c]{13cm}{
\texttt{\#\#\# Instruction:} \\ 
\texttt{\{new\_instruction\}} \\
\texttt{\#\#\# Response:} \\ 
\texttt{\{new\_response\}} 
}\\
\hline
\end{tabular}
\caption{Prompt template for training instruction-response pair expansion models. 
$N$ is randomly chosen from 1 to 3.}
\label{tab:train_exp_response_prompt}
\end{table*}

\subsection{Prompt Templates for Model Evaluation}

\begin{table*}[!h]
\small
\centering
\begin{tabular}{l|l}
\toprule
System prompt &
\parbox[c]{13cm}{
\texttt{You are a helpful and precise assistant for\\ checking the quality of the answer.}
} \\
\toprule
User prompt &
\parbox[c]{13cm}{
\texttt{[Instruction]} \\
\texttt{\{inst\}} \\ 
\texttt{[The Start of Assistant 1’s Answer]} \\ 
\texttt{\{ans1\}} \\ 
\texttt{[The End of Assistant 1’s Answer]} \\ 
\texttt{[The Start of Assistant 2’s Answer]} \\ 
\texttt{\{ans2\}} \\ 
\texttt{[The End of Assistant 2’s Answer]} \\
\texttt{[System]} \\
\texttt{We would like to request your feedback on the \textit{TRUTHFULNESS} of two AI assistants in response to the user instruction and input displayed above.}\\
\texttt{Please rate the \textit{TRUTHFULNESS} of their responses. Each assistant receives a \textit{TRUTHFULNESS} score on a scale of 1 to 10, where a higher score indicates better \textit{TRUTHFULNESS} performance.} \\
\texttt{Please first provide a comprehensive explanation of your evaluation, avoiding any potential bias and ensuring that the order in which the responses were presented does not affect your judgment. Then, output two lines indicating the scores for Assistant 1 and 2, respectively.} \\
\texttt{Output with the following format:} \\
\texttt{Evaluation evidence: <your evaluation explanation here>} \\
\texttt{Score of the Assistant 1: <score>} \\
\texttt{Score of the Assistant 2: <score>}
}\\
\bottomrule
\end{tabular}
\caption{Prompt template for evaluating the truthfulness of answers given by AI assistants.}
\label{tab:evaluation_prompt}
\end{table*}

\begin{table*}[!h]
\small
\centering
\begin{tabular}{l|l}
\toprule
System prompt &
\parbox[c]{13cm}{
\texttt{You are a helpful and precise assistant for\\ checking the quality of the answer.}
} \\
\toprule
User prompt &
\parbox[c]{13cm}{
\texttt{[Instruction]} \\
\texttt{\{inst\}} \\ 
\texttt{[The Start of Assistant 1’s Answer]} \\ 
\texttt{\{ans1\}} \\ 
\texttt{[The End of Assistant 1’s Answer]} \\ 
\texttt{[The Start of Assistant 2’s Answer]} \\ 
\texttt{\{ans2\}} \\ 
\texttt{[The End of Assistant 2’s Answer]} \\
\texttt{[System]} \\
\texttt{We would like to request your feedback on the \textit{LEVEL of DETAIL} of two AI assistants in response to the user instruction and input displayed above.}\\
\texttt{Please rate the \textit{LEVEL of DETAIL} of their responses. Each assistant receives a \textit{LEVEL of DETAIL} score on a scale of 1 to 10, where a higher score indicates better \textit{LEVEL of DETAIL} performance.} \\
\texttt{Please first provide a comprehensive explanation of your evaluation, avoiding any potential bias and ensuring that the order in which the responses were presented does not affect your judgment. Then, output two lines indicating the scores for Assistant 1 and 2, respectively.} \\
\texttt{Output with the following format:} \\
\texttt{Evaluation evidence: <your evaluation explanation here>} \\
\texttt{Score of the Assistant 1: <score>} \\
\texttt{Score of the Assistant 2: <score>}
}\\
\bottomrule
\end{tabular}
\caption{Prompt template for evaluating the level of detail of answers given by AI assistants.}
\label{tab:evaluation_prompt2}
\end{table*}

\end{document}